\documentclass[letterpaper]{article} % DO NOT CHANGE THIS
\usepackage[preprint]{aaai2027} 

\usepackage[hyphens]{url}
\usepackage{graphicx}
\usepackage{natbib}
\usepackage{caption}
\usepackage{algorithm}
\usepackage{algorithmic}

\usepackage{newfloat}
\usepackage{listings}
\DeclareCaptionStyle{ruled}{labelfont=normalfont,labelsep=colon,strut=off} % DO NOT CHANGE THIS
\floatstyle{ruled}
\newfloat{listing}{tb}{lst}{}
\floatname{listing}{Listing}

\usepackage{booktabs}

\usepackage{amsmath,amssymb,bm}
\usepackage{pifont,multirow}
\usepackage{amsthm}
\newtheorem{proposition}{Proposition}

\newcommand{\domain}{\Omega}

\newcommand{\Attn}{\mathrm{Attn}}
\newcommand{\PE}{\mathrm{PE}}

\newcommand{\model}{\textsc{TLO}}

\title{From Fixed Grids to Moving Particles:\\
A Transferable Latent Operator for Fluid Dynamics}
\author{
    Meng Li\textsuperscript{\rm 1,2},
    Chuqi Chen\textsuperscript{\rm 3},
    Zhengqing Gao\textsuperscript{\rm 1,4},
    Xi Zhou\textsuperscript{\rm 1},\\
    Xiao Sun\textsuperscript{\rm 1},
    Yang Xiang\textsuperscript{\rm 2,5}\corresponding,
    Huaxi Huang\textsuperscript{\rm 1}\corresponding
}
\affiliations{
    \textsuperscript{\rm 1}Shanghai Artificial Intelligence Laboratory,\\
    \textsuperscript{\rm 2}The Hong Kong University of Science and Technology,\\
    \textsuperscript{\rm 3}University of Michigan,\\
    \textsuperscript{\rm 4}Mohamed bin Zayed University of Artificial Intelligence,\\
    \textsuperscript{\rm 5}HKUST Shenzhen-Hong Kong Collaborative Innovation Research Institute\\
}

\begin{document}

\maketitle

\begin{abstract}
Lagrangian modeling is vital to fluid dynamics, as it characterizes particle transport and complements the Eulerian description. However, Lagrangian trajectories are less commonly available than Eulerian fields, while most neural operators are trained and evaluated primarily in the Eulerian representation. This mismatch motivates a new learning problem: can a model trained solely on Eulerian observations generalize zero-shot from Eulerian field prediction to Lagrangian particle rollout, without Lagrangian supervision or task-specific adaptation? To address this problem, we propose the \emph{Transferable Latent Operator} (\model{}), which learns a unified flow representation shared by Eulerian field prediction and Lagrangian particle rollout. \model{} decouples latent flow evolution from coordinate-dependent decoding: querying the evolving latent representation at fixed spatial coordinates yields Eulerian fields, whereas querying velocities at particle positions and recursively updating these positions enables Lagrangian rollout. Across five fluid-dynamics benchmarks, \model{} consistently outperforms existing neural operators in both Eulerian field prediction and zero-shot Lagrangian rollout, with further gains from limited Lagrangian fine-tuning.
\end{abstract}

\section{Introduction}

Lagrangian particle trajectories provide a direct description of transport,
mixing, and dispersion in fluid flows. However, most fluid-learning datasets
are available as time-dependent fields on fixed Eulerian grids
~\citep{lu2021deeponet,li2021fno,takamoto2022pdebench,ohana2024well}.
Native Lagrangian data are less commonly available because they require
specifying and tracking particle populations. At high particle counts,
trajectory generation, storage, and particle-based training can also become
expensive in time and memory
~\citep{sanchez2020gns,toshev2023lagrangebench,neuralmpm2024}.
This mismatch motivates a practical question: can a model trained only on
fixed-grid Eulerian data also support Lagrangian particle rollout of the same
flow?

We formulate this problem as \emph{zero-shot Eulerian-to-Lagrangian
generalization}. A model is trained exclusively with Eulerian field
supervision, without particle coordinates, particle velocities, or trajectory
labels. At test time, the same model is queried at moving particle positions,
and its predicted velocities are recursively integrated without parameter
adaptation. This is more than ordinary arbitrary-coordinate evaluation: each
velocity prediction changes the next particle position and therefore determines
the next query location. Errors can consequently induce a history-dependent
query shift and accumulate into trajectory drift.

% Existing methods connect Eulerian and Lagrangian representations through latent
% particles, joint field--trajectory learning, trajectory-based reconstruction,
% or coordinate-conditioned decoding
% ~\citep{trajectoryflownet2025,elpinn2025,li2022oformer,
% li2023gino,alkin2024upt,ma2024deeplag}. However, these methods use different forms of
% supervision or focus primarily on Eulerian prediction. A fixed-grid predictor
% combined with interpolation provides another natural baseline, but its particle
% rollout remains tied to the resolution and finite support of the output grid.
% Once particles leave this support, standard interpolation cannot continue
% without an additional extrapolation rule. We discuss these distinctions further
% in Related Work.
Although existing methods have explored Eulerian--Lagrangian modeling,
they do not directly address zero-shot Eulerian-to-Lagrangian rollout.
Approaches based on latent particles, joint field--trajectory learning, or
trajectory reconstruction
~\citep{trajectoryflownet2025,elpinn2025,alkin2024upt,ma2024deeplag}
typically require additional Lagrangian supervision or particle-specific
modeling, preventing zero-shot transfer. Alternatively, Eulerian neural operators~\cite{li2021fno,tripura2022wno,rahman2022uno} can be combined with numerical
interpolation to obtain particle velocities from predicted grid fields.
However, this strategy remains tied to the resolution and finite support of
the Eulerian discretization. Once particles move beyond the grid support,
standard interpolation cannot continue without additional extrapolation
rules. 

These limitations motivate a neural operator that learns flow dynamics
from Eulerian observations while supporting direct evaluation at recursively
evolving particle locations. We therefore introduce the \emph{Transferable Latent Operator}
(\model{}), which uses one latent flow evolution with two query modes. An
encoder maps fixed-grid Eulerian observations into spatially localized tokens
augmented with global flow context, and a shared latent processor advances this
representation independently of the output query set. A
coordinate-conditioned decoder then evaluates the predicted field at requested
locations. Querying on the fixed Eulerian grid produces an Eulerian field
forecast, whereas querying the velocity channels at recursively advected
particle positions produces a Lagrangian rollout. Particle coordinates are used
only as decoder queries and are not required by the Eulerian training
objective.

We evaluate \model{} on five fluid-dynamics benchmarks for Eulerian forecasting
and on three benchmarks for zero-shot Lagrangian particle rollout. \model{}
achieves strong Eulerian accuracy while transferring directly to reference-path
velocity prediction and closed-loop trajectories. We additionally compare with
a native particle-based model and grid interpolation. Direct decoding is
competitive with interpolation within the Eulerian support and remains
evaluable after particles leave that support, where standard interpolation
becomes undefined. We separately study decoder-only adaptation using sparse
particle-velocity supervision.

Our main contributions are summarized as follows:
\begin{itemize}
\item We formulate and evaluate \emph{zero-shot
Eulerian-to-Lagrangian generalization}, in which a model trained only with
fixed-grid Eulerian supervision performs closed-loop particle rollout without
particle supervision or parameter adaptation.

\item We propose the \emph{Transferable Latent Operator}, whose
output-query-independent latent evolution and coordinate-conditioned decoder
allow the same trained model to support both fixed-grid forecasting and moving
particle queries.

\item Across five Eulerian benchmarks and three particle-rollout benchmarks,
\model{} achieves strong performance, compares favorably with neural-operator
and native particle-based baselines, and remains evaluable beyond the fixed
Eulerian grid support.
\end{itemize}

\section{Related Work}
\label{sec:related_work}

\paragraph{Neural operators.}
Neural operators learn mappings between function spaces for PDE modeling
~\citep{lu2021deeponet,li2024harnessing,gao2025gaot}. Representative architectures include
spectral operators, coordinate-based models, and attention- or token-based
operators~\citep{li2021fno,hao2023gnot,li2023gino,wu2024transolver,alkin2024upt,cao2021gktrm}. Although several of these models support
arbitrary-coordinate evaluation, they are generally evaluated on fixed or
externally specified query locations. We instead consider moving queries whose
positions are recursively generated by the model's predicted velocities.

\paragraph{Eulerian--Lagrangian modeling.}
Classical particle-grid methods couple Eulerian fields and Lagrangian particles
through transfers between the two representations
~\citep{brackbill1986flip,sulsky1994mpm}. Recent neural methods use latent
particles to improve Eulerian forecasting~\citep{ma2024deeplag}, infer fields or
trajectories from particle observations
~\citep{trajectoryflownet2025,elpinn2025}, or learn particle-based simulators
~\citep{neuralmpm2024,neuralvortex2020}. In contrast, the zero-shot version of
\model{} is trained only on fixed-grid Eulerian fields, with particle positions
introduced at inference time solely as decoder queries. The same learned flow
representation therefore supports both Eulerian field prediction and
closed-loop Lagrangian rollout. Sparse decoder adaptation is studied separately
as a non-zero-shot extension.

\section{Problem Setting}\label{sec:problem_setting}

Let $\Omega\subset\mathbb{R}^d$ be the spatial domain. At time $t$, the
state of a flow is represented by a vector-valued field
$\bm{u}_t:\Omega\rightarrow\mathbb{R}^{d_u}$, where the $c$ channels may contain
velocity, pressure, density, or other physical variables. We denote the
velocity by
$\bm{v}_t(\bm{x}):=\Pi_v\bm{u}_t(\bm{x})\in\mathbb{R}^d$, where $\Pi_v$ is the projection operator extracting the velocity channels.

The available data do not provide the continuous function $\bm{u}_t$
directly. Instead, each state is observed at a fixed set of Eulerian
coordinates
\begin{equation}
\begin{aligned}
    X_E:=\{\bm{x}_i\}_{i=1}^{N_E},\,
    \mathbf{U}_t:=[\bm{u}_t(\bm{x}_i)]_{i=1}^{N_E}
    \in\mathbb{R}^{N_E\times d_u}.
\end{aligned}
    \label{eq:eulerian_observation}
\end{equation}
The coordinates in $X_E$ remain fixed over time, while the field values
$\mathbf{U}_t$ evolve. The dataset therefore consists of temporal sequences
$\{\mathbf{U}_0,\mathbf{U}_1,\ldots\}$ observed on the same Eulerian coordinates.

Our goal is to learn the one-step evolution from the current Eulerian
observation $\mathbf{U}_t$ to the next physical field
$\bm{u}_{t+1}$. In particular, the learned operator should not be restricted
to returning values only on $X_E$. Given any finite set of output coordinates
$Q=\{\bm{q}_j\}_{j=1}^{N_Q}\subset\Omega$, it predicts
\[
\begin{aligned}
    \widehat{\mathbf{U}}_{t+1}(Q):=\mathcal{G}_\theta(X_E,\mathbf{U}_t;Q)
    &=
    [\widehat{\bm{u}}_{t+1}(\bm{q}_1),\ldots,
     \widehat{\bm{u}}_{t+1}(\bm{q}_{N_Q})]^\top .
\end{aligned}
\]
Here, a \emph{query} is simply a coordinate $\bm{q}_j$ at which the
next field is requested. Setting $Q=X_E$ produces the next Eulerian grid
field. Setting $Q$ to particle coordinates provides the local velocity values
needed to evolve Lagrangian particles.

\begin{figure}[!htbp]
    \centering
\includegraphics[width=1\linewidth]{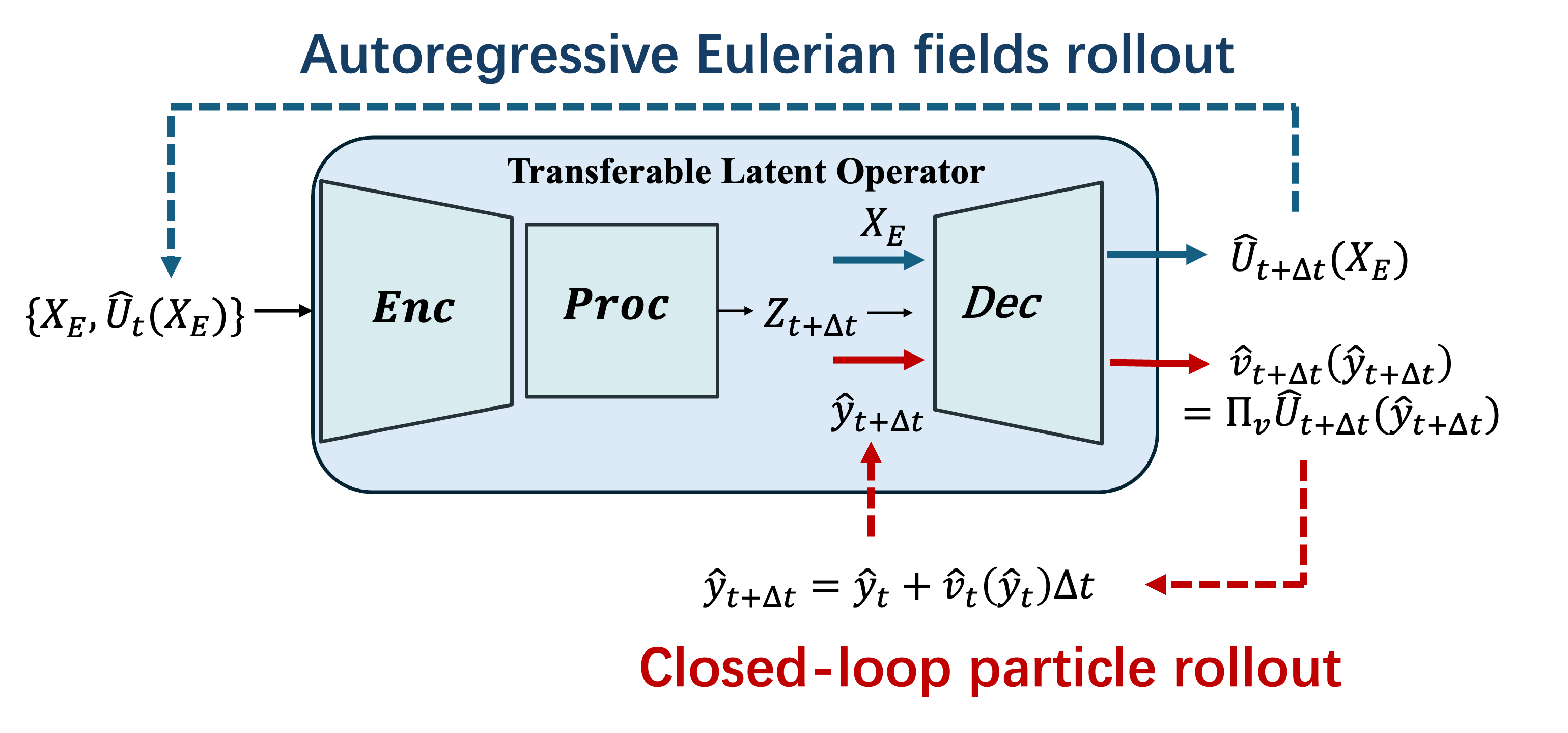}
\caption{Workflow of \model{} for unified Eulerian field forecasting and
Lagrangian particle rollout from a shared latent flow representation. Notations $X_E$, $\hat{U}_t$, and $\hat{y}_t$ follow Section Problem Setting.}
\end{figure}

\paragraph{Learning from Eulerian fields.}
Fluid-dynamics datasets are commonly available as time-dependent fields on
fixed spatial grids. We therefore use consecutive Eulerian fields as the
primary source of supervision. Although the operator accepts a general query
set $Q$, its training targets are provided only on $X_E$:
\begin{equation}
    \mathcal{L}_{\mathrm{Eul}}
    =
    \mathbb{E}_t\!\left[
    \frac{1}{N_E  d_u}
    \left\|
        \mathcal{G}_\theta(X_E,\mathbf{U}_t;X_E)
        -\mathbf{U}_{t+1}
    \right\|_F^2
    \right].
    \label{eq:eulerian_training_objective}
\end{equation}
Thus, neither particle coordinates nor trajectory labels are required to
train the field-evolution model. Starting from an observed initial state
$\mathbf{U}_0$, choosing $Q=X_E$ at every step gives the autoregressive
Eulerian rollout
\begin{equation}
    \widehat{\mathbf{U}}_{t+1}
    =
    \mathcal{G}_\theta(
        X_E,\widehat{\mathbf{U}}_t;X_E
    ),
    \qquad
    \widehat{\mathbf{U}}_0=\mathbf{U}_0.
    \label{eq:eulerian_rollout}
\end{equation}
This produces the predicted field sequence
$\{\widehat{\mathbf{U}}_1,\ldots,\widehat{\mathbf{U}}_H\}$ on the
fixed Eulerian grid support. Besides providing a field forecast, this rollout supplies
the evolving flow state from which velocities at moving coordinates can be
evaluated.

\begin{figure*}[t]
    \centering
    \includegraphics[width=\linewidth]{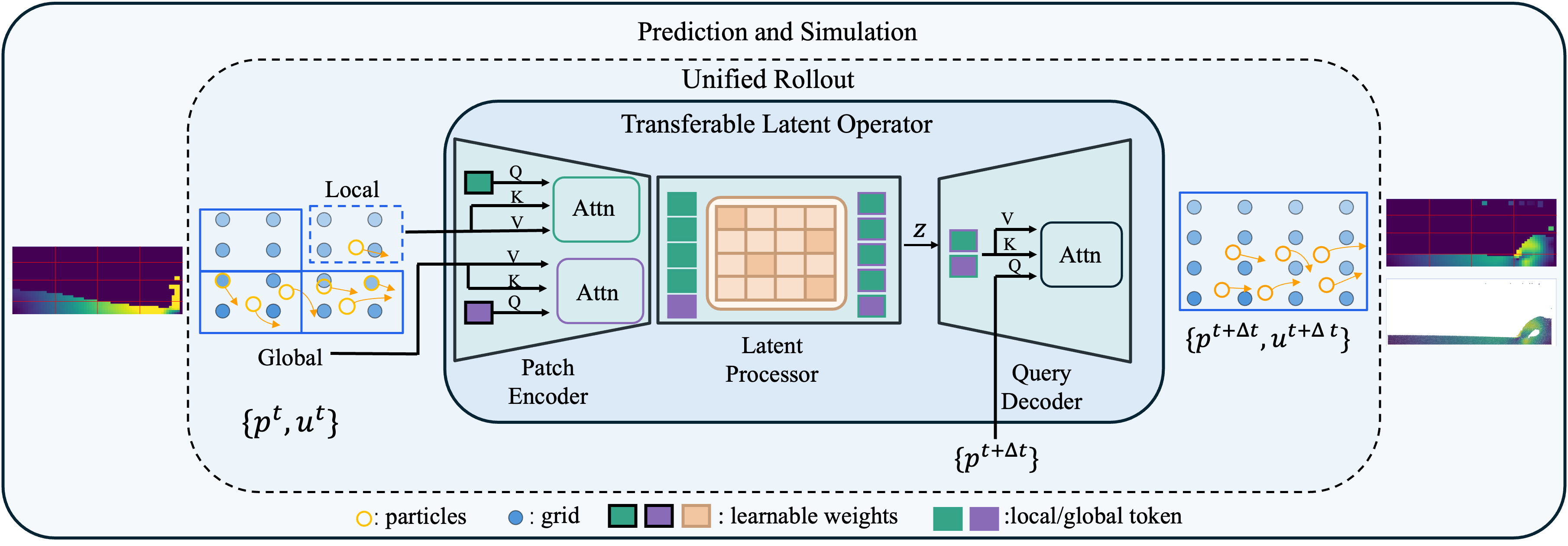} 
    \caption{ Overview of the Transferable Latent Operator (\model{}).
TLO learns coordinate-independent latent dynamics from Eulerian observations.
The hierarchical encoder extracts local and global flow representations, while
the latent processor evolves the fused tokens without spatial coordinates. A
coordinate-conditioned decoder then queries the latent state at arbitrary
locations, enabling both fixed-grid Eulerian prediction and Lagrangian rollout.
Particle positions are used only as inference-time queries and are not required
for Eulerian training.}
    \label{fig:ours_arch}
\end{figure*}

\paragraph{Recovering Lagrangian trajectories.}
Fixed-grid forecasts describe how the flow evolves at stationary spatial
locations. Many transport problems, however, concern material carried by the
flow and therefore require following moving particles. Particle trajectories
provide a direct description of transport and expose accumulated drift that
may not be evident from an average fixed-grid field error. To obtain this
Lagrangian view, let
$Y_0=\{\bm{y}_{i,0}\}_{i=1}^{N_P}$ denote a set of initial particle
positions. Once $\widehat{\bm{u}}_t$ has been predicted, its velocity channels
are evaluated at the current particle positions and integrated through
\begin{equation}
    \widehat{\bm{y}}_{i,t+1}
    =\mathcal{B}\!\left(
        \widehat{\bm{y}}_{i,t}
        +\Delta t\,
        \widehat{\bm{v}}_{t}(\widehat{\bm{y}}_{i,t})
      \right),
    \qquad
\widehat{\bm{y}}_{i,0}=\bm{y}_{i,0},
    \label{eq:closed_loop_pathline}
\end{equation}
where $\mathcal B$ enforces the physical boundary condition during particle
integration, ensuring that the updated particle positions remain within the
spatial domain $\Omega$. The Eulerian
state continues to evolve through Equation~\eqref{eq:eulerian_rollout}; the
particles only determine where the predicted velocity field is evaluated. Thus, in the Eulerian-only setting, the same trained operator predicts fields
on the fixed grid and provides velocities at moving particle positions. The
particle trajectories are then obtained by numerical integration, without
training a separate particle-dynamics model.

\paragraph{Field and trajectory errors.}
\label{para:eval_metric}

Since Lagrangian rollout is driven by velocities queried from the predicted
flow, fixed-grid field error alone does not fully characterize model
performance. We therefore distinguish errors in Eulerian field prediction,
velocity evaluation at particle positions, and the resulting closed-loop
trajectories. We refer to these quantities as \emph{Eul}, \emph{Ref}, and \emph{Path},
respectively, and report them throughout the experiments.

Let $H$ denote the rollout horizon, $N_E$ the number of Eulerian
grid points, and $N_p$ the number of particles. We use $d_u$, $d_v$, and $d_x$ for the dimensions of
the predicted state, velocity, and particle position, respectively.

First, the fixed-grid Eulerian error is
\begin{equation}
    \mathcal E_{\mathrm{Eul}}
    :=
    \frac{1}{H N_E d_u}
    \sum_{t=1}^{H}
    \sum_{\bm g\in X_E}
    \left\|
        \widehat{\bm u}_t(\bm g)
        -
        \bm u_t(\bm g)
    \right\|_2^2 .
    \label{eq:eulerian_field_metric}
\end{equation}
This measures the autoregressive field prediction on the Eulerian support
$X_E$.

% To evaluate velocity without particle-position feedback, we use the reference
% particle positions $\bm y_{i,t}$ as query coordinates:
To evaluate velocity without position feedback, we query at the reference
positions $\bm y_{i,t}$:
\begin{equation}
    \mathcal E_{\mathrm{Ref}}
    :=
    \frac{1}{H N_p d_v}
    \sum_{t=1}^{H}
    \sum_{i=1}^{N_p}
    \left\|
        \widehat{\bm v}_t(\bm y_{i,t})
        -
        \bm v_t(\bm y_{i,t})
    \right\|_2^2 .
    \label{eq:reference_velocity_metric}
\end{equation}
The Eulerian field is still predicted autoregressively, but the query
coordinates are reset to the reference pathline at every step. Hence,
$\mathcal E_{\mathrm{Ref}}$ measures velocity accuracy along the correct
trajectory without propagating particle-position errors.

Finally, in the closed-loop rollout, the particle positions
$\widehat{\bm y}_{i,t}$ are recursively generated using the model's queried
velocities. The resulting trajectory error is
\begin{equation}
    \mathcal E_{\mathrm{Path}}
    :=
    \frac{1}{H N_p d_x}
    \sum_{t=1}^{H}
    \sum_{i=1}^{N_p}
    d_\Omega^2\!\left(
        \widehat{\bm y}_{i,t},
        \bm y_{i,t}
    \right),
    \label{eq:closed_loop_position_metric}
\end{equation}
where $d_\Omega$ accounts for the domain geometry and uses the minimum-image
distance on periodic domains.

\section{Transferable Latent Operator}\label{sec:method}

\model{} follows an encode--process--decode architecture~\cite{sanchez2020gns} that separates latent
flow modeling from coordinate-dependent field evaluation. Given an Eulerian
observation on
$X_E=\{\bm x_i\}_{i=1}^{N_E}$, we denote its sampled state by
$\mathbf U_t=[\bm u_t(\bm x_i)]_{i=1}^{N_E}$. The model advances this state
by one time step and evaluates the predicted field at a specified query set
$Q=\{\bm q_j\}_{j=1}^{N_Q}$.

Formally, the one-step operator $\mathcal G_\theta$ is decomposed into an
encoder $\mathcal E_\theta$, a latent processor $\mathcal T_\theta$, and a
coordinate-conditioned decoder $\mathcal D_\theta$:
\begin{equation}
\begin{aligned}
    \bm Z_t^{0}
    &=
    \mathcal E_\theta(X_E,\mathbf U_t),\\
    \bm Z_t^{L}
    &=
    \mathcal T_\theta(\bm Z_t^{0}),\\
    \widehat{\mathbf U}_{t+1}\big|_{Q}
    &=
    \mathcal D_\theta(\bm Z_t^{L},Q).
\end{aligned}
\label{eq:tlo_factorization}
\end{equation}
The encoder constructs a latent representation of the observed flow, the
processor advances this representation, and the decoder evaluates the
predicted state at the requested coordinates. Importantly, the encoder and
latent processor are independent of $Q$; only the final readout depends on the
output coordinates. Setting $Q=X_E$ produces the next Eulerian field, whereas
setting $Q$ to the current particle positions provides the velocities used for
Lagrangian rollout. The following sections describe each component in detail.

\textbf{Patch Encoder} employs cross-attention as a permutation-invariant
aggregation mechanism to map local and global  observations into a fixed-size latent representation. 

To capture both localized transport dynamics and long-range flow dependencies,
we construct a hierarchical domain representation consisting of local patches
$\domain_p$ and the global domain $\domain_g$:
\[
\mathcal{D}
=
\{
\underbrace{\domain_{1},\dots,\domain_{P}}_{\text{local patches}},
\underbrace{\domain_g}_{\text{global context}}
\},
\qquad
\domain_g=\bigcup_{p=1}^{P}\domain_p .
\]

For each region $\domain_*\in\mathcal{D}$, we first lift coordinate-attached
observations into position-physical features. Specifically,
\[
\begin{aligned}
\bm r_i
&=
\phi
\left[
\PE(\bm x_i),
\PE(\bm x_i^{\,p}),
\bm u_i,
e_p
\right],\\
R_*
&=
\{\bm r_i \mid \bm x_i\in\domain_*\},
\end{aligned}
\]
where $\bm x_i$ denotes the global coordinate in $\domain$, 
$\bm x_i^{\,p}$ represents the local coordinate relative to its patch,
and $\bm u_i$ is the associated physical state. 
The fixed Fourier feature encoding $\PE(\cdot)$ provides continuous coordinate
information, while the learnable patch embedding $e_p$ identifies local
spatial regions. The lifting function $\phi$ follows the standard formulation
used in neural operator learning~\cite{li2021fno}.

For each local patch and the global domain, learnable latent queries
$\bm Q^l$ and $\bm Q^g$ aggregate observations through cross-attention:
\[
\bm Z_*
=
\Attn
(
\bm Q^*,
K(R_*),
V(R_*)
),
\]
where $*$ indicates either a local patch or the global domain. 
The resulting latent tokens from different spatial regions are fused to form
the initial latent state:
\[
\bm Z^0
=
\text{concat}
(\{\bm Z_*\})
\]
By integrating local coordinate encoding with global context aggregation,
the hierarchical Patch Encoder converts coordinate-attached
observations into a discretization-independent latent representation.
This latent state preserves both local transport structures and global flow
dependencies, providing a unified representation that can be queried on fixed
Eulerian grids or evolving Lagrangian particle trajectories.

\textbf{The Latent Processor} 
learns the temporal evolution of underlying dynamics and separats physical evolution from spatial discretization. We implement the Latent Processor with $L$ attention-based blocks inspired by
the slice-based aggregation mechanism in Transolver~\citep{wu2024transolver}.
Different from coordinate-conditioned neural operators, the proposed processor
operates purely on latent tokens.

At layer $\ell$, the latent tokens are flattened as
$\bm Z^\ell\in\mathbb{R}^{N\times h}$, where $N$ denotes the total number of
latent tokens. The processor first assigns tokens into $M$ adaptive latent
slices:
\[
\bm W^\ell
=
\operatorname{softmax}(\psi(\bm Z^\ell))
\in\mathbb{R}^{N\times M},
\]
where each row of $\bm W^\ell$ represents the contribution of a token to the
latent slices. The slice representations are obtained by weighted and are then
updated through self-attention:
\[
\bm S^\ell
=
\operatorname{Attn}(\hat{\bm S}^{\ell}), \,
\hat{\bm S}^{\ell}
=
\widetilde{\bm W}^{\ell,T}\bm Z^\ell,
\]
where $\widetilde{W}^{\ell}_{i,j} = W^{\ell}_{i,j}/\sum_i W^{\ell}_{i,j}$. The updated slice information is distributed back to the latent tokens:
\[
\bm Z^{\ell+1}
=
\bm Z^\ell
+
\operatorname{Proj}
(\bm W^\ell\bm S^\ell)
+
\alpha\operatorname{FFN}(\bm Z^\ell).
\]

Through latent-space evolution independent of spatial coordinates, the processor
provides a reusable dynamic representation $Z^L$ that can be decoded at arbitrary
Eulerian locations or evolving Lagrangian particle positions.

\textbf{Query Decoder} provides a coordinate-conditioned readout from the
coordinate-independent latent state, decoupling latent dynamics evolution from
spatial discretization. For each query point, we first identify its corresponding patch
$\domain_{p(j)}$ and construct a query representation
\[
\bm g_j
=
\psi[
\PE(\bm q_j),
\PE(\bm q_j^{\,p})
],
\]
using the previous embedding strategy. The physical quantity at the query location is recovered through
cross-attention between the query feature and latent tokens from neighboring
patches:
\[
\hat{\bm u}(\bm q_j)
=
\sum_{p_k\in\mathcal N(\bm q_j)}
\alpha_k
\operatorname{Attn}
\left(
Q(\bm g_j),
K(\bm Z_{p_k}),
V(\bm Z_{p_k})
\right),
\]
where $\mathcal N(\bm q_j)$ denotes the set of neighboring patches
around the patch containing $\bm q_j$, and $\alpha_k$ represents the
aggregation weight of each neighboring patch.

\section{Error Analysis}
\label{sec:closed_loop_error}

\model{} predicts velocity, whereas a Lagrangian rollout is ultimately
evaluated through particle position. Consider particle $i$ over a rollout of
$T$ steps, indexed by $t=0,\ldots,T$. Let $\Delta t$ be the integration step,
and let $\bm v_t$ and $\widehat{\bm v}_t$ denote the reference and predicted
velocity fields at step $t$, respectively. The corresponding reference and
predicted particle positions are denoted by $\bm y_{i,t}$ and
$\widehat{\bm y}_{i,t}$.

Using forward Euler, their positions evolve as
\begin{equation}
\begin{aligned}
    \bm y_{i,t+1}
    &=
    \mathcal B\!\left(
        \bm y_{i,t}
        +\Delta t\,\bm v_t(\bm y_{i,t})
    \right),\\
    \widehat{\bm y}_{i,t+1}
    &=
    \mathcal B\!\left(
        \widehat{\bm y}_{i,t}
        +\Delta t\,\widehat{\bm v}_t(\widehat{\bm y}_{i,t})
    \right),
    \, t=0,\ldots,T-1.
\end{aligned}
\label{eq:error_analysis_updates}
\end{equation}
where $\mathcal B$ is the bounce operator that handles particles that reach the boundary of the physical-domain 
during rollout. Define the position error and the velocity error evaluated at
the reference particle position as
\begin{equation}
    \delta_{i,t}
    :=
    d_\Omega(\widehat{\bm y}_{i,t},\bm y_{i,t}),\,
    r_{i,t}
    :=
    \left\|
        \widehat{\bm v}_t(\bm y_{i,t})
        -
        \bm v_t(\bm y_{i,t})
    \right\|_2 .
\label{eq:error_analysis_definitions}
\end{equation}

\begin{proposition}[Velocity error to trajectory error]
\label{prop:velocity_to_trajectory}
Assume that $\widehat{\bm y}_{i,0}=\bm y_{i,0}$, that $\mathcal B$ is
non-expansive under $d_\Omega$, and that $\widehat{\bm v}_t$ is
$\widehat L_t$-Lipschitz over the region visited by the two particles. Then,
for every $t=0,\ldots,T-1$,
\begin{equation}
    \delta_{i,t+1}
    \leq
    (1+\Delta t\,\widehat L_t)\delta_{i,t}
    +
    \Delta t\,r_{i,t}.
\label{eq:velocity_to_trajectory_recursion}
\end{equation}
Consequently, the position error at the rollout horizon $T$ satisfies
\begin{equation}
    \delta_{i,T}
    \leq
    \Delta t
    \sum_{s=0}^{T-1}
        r_{i,s}
        \prod_{k=s+1}^{T-1}
        (1+\Delta t\,\widehat L_k),
\label{eq:velocity_to_trajectory_bound}
\end{equation}
with the convention that an empty product equals one.
\end{proposition}

Proposition~\ref{prop:velocity_to_trajectory} connects the velocity predicted
by the model to the particle position evaluated in a Lagrangian rollout. At
each step, a velocity error $r_{i,t}$ causes an immediate position error of
approximately $\Delta t\,r_{i,t}$. The updated position is then used as the
query coordinate at the next step. Therefore, once the predicted particle
deviates from the reference path, subsequent velocities are evaluated at a
different location, which may introduce further error. The factor
$1+\Delta t\,\widehat L_t$ describes the strength of this feedback. Over a
long rollout, these errors accumulate, and errors made earlier can affect more
subsequent steps. Thus, a small \emph{Ref vel.} error is important but does
not necessarily imply a small \emph{Path} error, which motivates reporting
both metrics.

\section{Experiments}
\label{sec:experiments}

\paragraph{Benchmarks.}
We evaluate \model{} on five fluid benchmarks covering particle-based
simulations, grid-based PDEs, and real-world SEA (ocean current). DAM2D and
TGV3D are SPH datasets from LagrangeBench~\citep{toshev2023lagrangebench}
and are converted to Eulerian fields by kernel splatting; see supplementary material for details. NS2D,
Burgers3D, and SEA provide grid-based velocity fields from,
respectively, incompressible Navier--Stokes dynamics, nonlinear
advection--diffusion, and ocean reanalysis
data~\citep{kovachki2023neural,koehler2024apebench,ma2024deeplag}.
DAM2D uses its native particle trajectories, while virtual tracers for
Burgers3D and SEA are generated from the reference velocity
fields using Parcels~\citep{delandmeter2019parcels}. These particles are
passive tracers and do not influence the Eulerian dynamics.

\begin{table}[!ht]
    \centering
    \label{tab:benchmarks}
    \small
    \setlength{\tabcolsep}{3pt}
    \begin{tabular}{lcccc}
    \toprule
    Dataset & Native data & Support & Evaluation & Horizon \\
    \midrule
    DAM2D & SPH & $32\times64$ & Eul/Ref/Path & 10 \\
    NS2D & Grid & $64^2$ & Eul & 10 \\
    TGV3D & SPH & $20^3$ & Eul & 5 \\
    Burgers3D & Grid & $32^3$ & Eul/Ref/Path & 5 \\
    SEA & Grid & $180\times300$ & Eul/Ref/Path & 10 \\
    \bottomrule
\end{tabular}
\caption{Benchmark summary. Eul, Ref, and Path denote fixed-grid field,
    reference-query velocity, and closed-loop trajectory evaluation.}
\end{table}

\paragraph{Baselines.}
We compare with LSM~\citep{wu2023LSM}, GINO~\citep{li2023gino}, GNOT~\citep{hao2023gnot},
Transolver~\citep{wu2024transolver}, UPT~\citep{alkin2024upt},
LNO~\citep{wang2024lno}, and DeepLag~\citep{ma2024deeplag}, covering latent-space, geometry-aware, attention-based, and
Eulerian--Lagrangian operator architectures.
For particle evaluation, query-native models are evaluated directly at
particle coordinates, whereas grid-output models use bilinear or trilinear
interpolation of their predicted fields. We additionally evaluate both direct
decoding and grid interpolation from the same \model{} checkpoint.

\paragraph{Training and evaluation.}
All models are trained for one-step Eulerian field prediction and evaluated
through autoregressive rollout. We report fixed-grid field error
(\emph{Eul}), velocity error along reference particle trajectories
(\emph{Ref}), and closed-loop particle-position error (\emph{Path}), as
defined in Eqs. (5)–(7). For Ref, reference particle
positions are supplied as queries at each step, while the predicted field
continues to evolve autoregressively. For Path, only the initial particle
positions are given; subsequent positions are obtained by integrating the
predicted velocities and are used as the next queries. Thus, Path measures the
accumulated effect of velocity errors on the particle trajectory. The main
experiments use only Eulerian training data, with the sparse decoder-adaptation
study reported separately.

\subsection{Eulerian prediction}

\paragraph{Autoregressive Eulerian field rollout.} We first examine whether
\model{} preserves predictive accuracy in the standard Eulerian setting.
Starting from the ground-truth initial field, each model recursively predicts
the next state on the fixed Eulerian grid and feeds its previous prediction
back as input. We report the  fixed-grid field error (\emph{Eul}) defined in Eq. (5), averaged over all rollout steps $H$, spatial
locations, state channels, and test trajectories. 
This evaluation isolates temporal field-forecasting error from spatial-query
error and particle-position feedback.

As shown in Table~\ref{tab:main_results}, \model{} achieves strong Eulerian
rollout performance across all five benchmarks. The improvements cover SPH-derived and
grid-native data, two- and three-dimensional systems, and both simulated
and reanalysis flows. These results show that \model{} retains strong fixed-grid Eulerian forecasting accuracy while supporting moving particle queries through the same learned representation.

\begin{table}[!ht]
\centering
\footnotesize
\setlength{\tabcolsep}{3pt}
\renewcommand{\arraystretch}{1.1}
\begin{tabular}{@{}lccccc@{}}
\toprule
Method
& DAM2D
& NS2D
& TGV3D
& Burgers3D
& SEA \\
\midrule
LSM
    & 2.97E-02
    & 3.46E-03
    & 9.57E-04
    & 2.17E-03
    & 5.72E-01 \\
GINO
    & 6.04E-03
    & 3.68E-05
    & 2.47E-04
    & 5.35E-03
    & 5.90E-01 \\
GNOT
    & 2.24E-02
    & 1.72E-03
    & 9.37E-04
    & 4.21E-04
    & 4.19E-01 \\
Transolver
    & 6.13E-03
    & 1.40E-04
    & 7.67E-04
    & 4.22E-04
    & 5.22E-01 \\
LNO
    & 3.75E-03
    & 8.73E-05
    & 7.54E-03
    & 4.42E-03
    & 5.21E-01 \\
UPT
    & 2.38E-02
    & 1.31E-03
    & 7.54E-03
    & 5.59E-03
    & 8.97E-01 \\
DeepLag
    & 1.65E-02
    & 4.04E-03
    & 3.17E-04
    & 3.38E-03
    & 8.17E-01 \\
\midrule
\textbf{\model{} (Ours)}
    & \textbf{3.12E-03}
    & \textbf{1.98E-05}
    & \textbf{1.75E-04}
    & \textbf{9.88E-05}
    & \textbf{1.77E-01} \\
\bottomrule
\end{tabular}
\caption{Mean squared error (MSE) of autoregressive Eulerian rollout. Lower is better. The best result in each column is highlighted in bold.}\label{tab:main_results}
\end{table}

\begin{figure}[!htp]
    \centering
    \includegraphics[width=\linewidth]{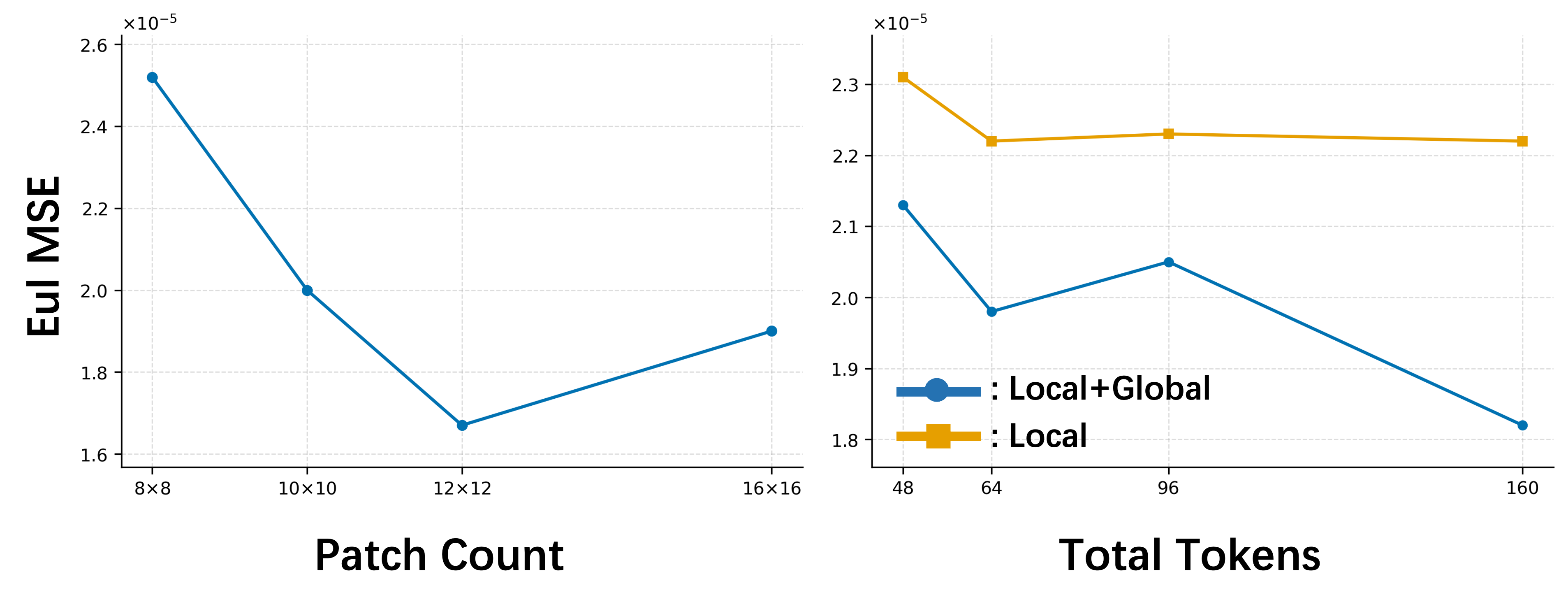}
    \caption{Patch and token ablations on NS2D. Left: Eul MSE versus patch
lattice size. Right: two models use matched total tokens, with a
fixed \(8\times8\) patch and global token equals 8 for
Local+Global. Lower is better.}
    \label{fig:ablation_local_global}
\end{figure}

\paragraph{Model ablations analysis.}
Figure~\ref{fig:ablation_local_global} shows that the \(12\times12\) patch
lattice yields the lowest error, suggesting a trade-off
between finer spatial partitioning and the amount of information available
within each patch. 

Under the same total token budget, Local+Global consistently
outperforms Local, showing that its gain
cannot be explained by additional latent capacity alone. And the Local+Global error exhibits an overall downward
trend and reaches its lowest value at 160 tokens. These results support the
complementary roles of local tokens for spatially resolved information and
global tokens for domain-level context.

In Table~\ref{tab:ablation_modules}, all variants are controlled with comparable model sizes to ensure a fair comparison. TLO has 150.8K parameters versus 146.7K without FFN, while Local-only and
Global-only use the same number of tokens as TLO. Thus, the gains cannot be
explained solely by model capacity, but arise from the local--global
representation and latent processing.

\begin{table}[t]
\centering
\begin{tabular}{lc}
\toprule
Design & Eul $\downarrow$ \\
\midrule
TLO (Local 56 + Global 8) & \textbf{1.98E-05} \\
Local-only (64) & 2.22E-05 \\
Global-only (64) & 4.21E-05 \\
w/o FFN & 4.51E-05 \\
\bottomrule
\end{tabular}
\caption{Ablation of TLO components on NS2D under comparable model sizes.
Token-allocation variants share the same 64-token budget. Lower L2 error is better.}\label{tab:ablation_modules}
\end{table}

\subsection{Transfer from Eulerian to Lagrangian rollout}\label{sec:e2l}

Having established strong Eulerian forecasting accuracy, we next evaluate
whether the learned dynamics can transfer from fixed-grid field prediction to
closed-loop rollout at moving particle coordinates. Because the latent
processor in \model{} evolves the flow representation independently of the
output query set, the same Eulerian-trained checkpoint can be used by changing
the decoder queries from fixed-grid locations to recursively advected particle
positions.

We first consider the \emph{Eulerian-only} (EO) setting. The model is trained
using only Eulerian field supervision, without particle coordinates, particle
velocities, or trajectory labels during training. At inference time, particle
positions are introduced only as decoder queries, and the queried velocities
are recursively integrated to form a closed-loop rollout. As shown in
Table~\ref{tab:e2l}, \model{} achieves the lowest Ref and Path errors among the
compared methods on all three Lagrangian benchmarks. These results show that an
Eulerian-trained \model{} checkpoint can transfer directly to moving-particle
queries without particle-based training supervision or parameter adaptation.
They also motivate evaluating closed-loop trajectories in addition to
fixed-grid forecasting accuracy.

We further consider a \emph{decoder-adapted} (DA) setting. Starting from the
Eulerian-trained checkpoint, we fine-tune only the decoder using sparse velocity
samples at particle coordinates, while keeping the encoder and latent processor
fixed. This adaptation provides additional, dataset-dependent improvements in
the coordinate readout. The results are consistent with the Eulerian-trained
latent representation already containing transport-relevant information, which
can be further calibrated for moving-coordinate evaluation through limited
particle supervision.

\begin{figure}[t]
    \centering
\includegraphics[width=\linewidth]{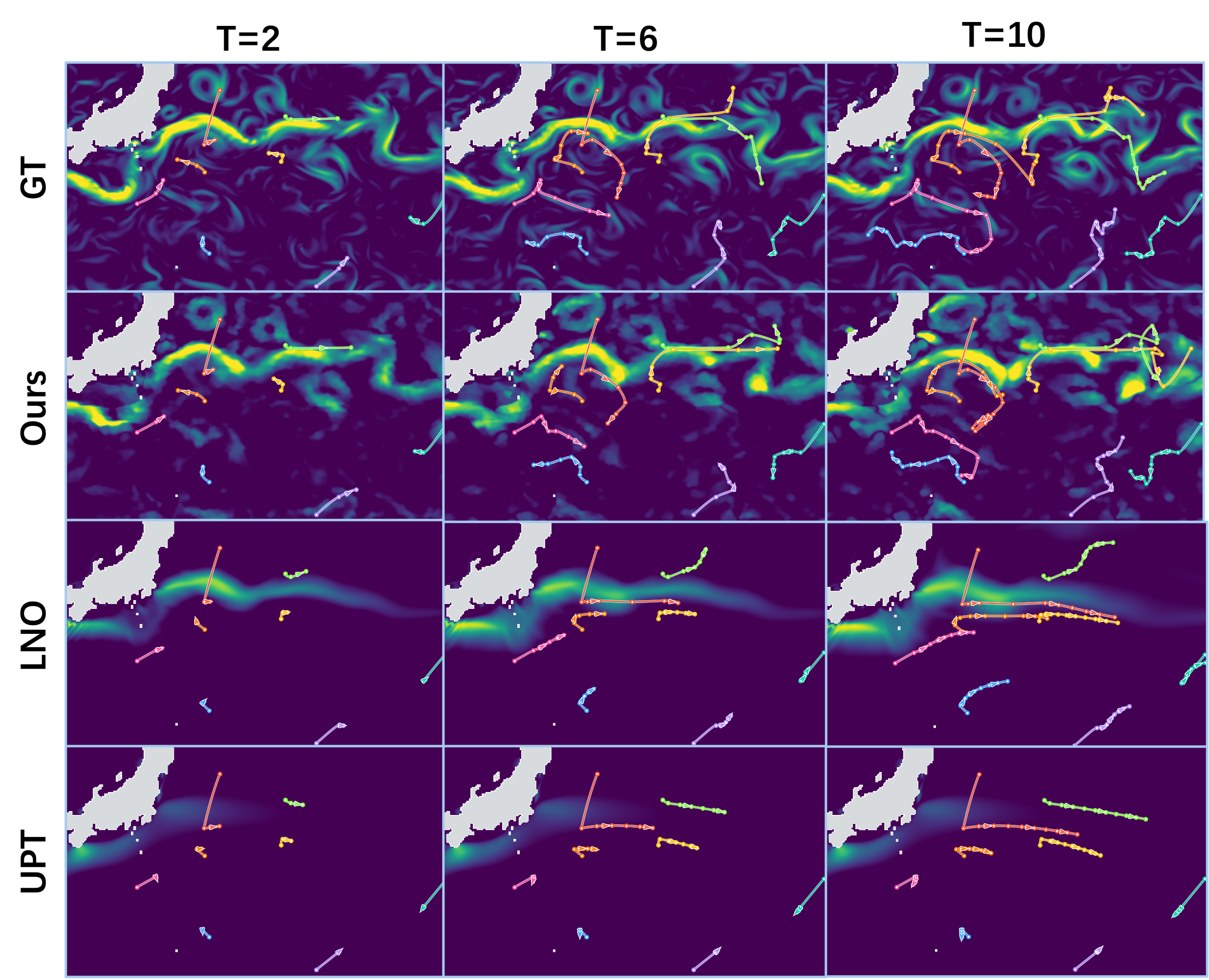}
    \caption{Qualitative closed-loop Lagrangian rollout on SEA.
Each column shows the Eulerian velocity field and the corresponding particle
trajectories at different rollout steps ($T=2,6,10$).}
    \label{fig:sea_rollout}
\end{figure}

Figure~\ref{fig:sea_rollout} provides a qualitative comparison of closed-loop
particle rollouts. All methods start from the same initial particle positions
and recursively update them using their predicted velocities. The resulting
trajectory deviations visualize the accumulated effects of velocity-prediction
error and feedback through recursively generated query locations.

\begin{table}[!ht]
\centering
\begingroup
\footnotesize
\setlength{\tabcolsep}{2pt}
\begin{tabular}{@{}lclccc@{}}
\toprule
Dataset & Setting & Method & Eul $\downarrow$ & Ref $\downarrow$ & Path $\downarrow$ \\
\midrule
\multirow{6}{*}{DAM2D}
    & \multirow{3}{*}{EO}
    & UPT
    & 2.38E-02 & 7.62E-02 & 1.44E-01 \\
    & & LNO
    & 3.75E-03 & 4.35E-03 & 5.30E-03 \\
    & & \textbf{\model{} (Ours)}
    & \textbf{3.12E-03} & \textbf{2.41E-03} & \textbf{3.72E-03} \\
\cmidrule(lr){2-6}
    & {DA} & \textbf{\model{} (Ours)}
    & \textbf{3.08E-03} & \textbf{2.39E-03} & \textbf{3.72E-03} \\
\specialrule{0.4pt}{1pt}{0pt}
\specialrule{0.4pt}{0.7pt}{1pt}
\multirow{6}{*}{Burgers3D}
    & \multirow{3}{*}{EO}
    & UPT
    & 5.59E-03 & 2.91E-03 & 7.73E-03 \\
    & & LNO
    & 4.42E-03 & 3.36E-03 & 9.42E-03 \\
    & & \textbf{\model{} (Ours)}
    & \textbf{9.88E-05} & \textbf{6.17E-04} & \textbf{3.01E-03} \\
\cmidrule(lr){2-6}
    &{DA} & \textbf{\model{} (Ours)}
    & \textbf{9.48E-05} & \textbf{2.25E-04} & \textbf{2.59E-03} \\
\specialrule{0.4pt}{1pt}{0pt}
\specialrule{0.4pt}{0.7pt}{1pt}
\multirow{6}{*}{SEA}
    & \multirow{3}{*}{EO}
    & UPT
    & 8.97E-01 & 4.92E-02 & 1.06E-02 \\
    & & LNO
    & 5.21E-01 & 5.29E-02 & 1.10E-02 \\
    & & \textbf{\model{} (Ours)}
    & \textbf{1.77E-01} & \textbf{3.00E-02} & \textbf{8.99E-03} \\
\cmidrule(lr){2-6}
    &{DA} & \textbf{\model{} (Ours)}
    & \textbf{1.72E-01} & \textbf{2.95E-02} & \textbf{8.29E-03} \\
\bottomrule
\end{tabular}
\endgroup
\caption{Transfer from Eulerian training to Lagrangian rollout. EO directly applies the Eulerian-trained checkpoint without particle-based training supervision. For \model{}, DA fine-tunes only the decoder using sparse velocity samples at particle coordinates. Eul, Ref, and Path denote fixed-grid field, reference-path velocity, and closed-loop particle-position MSE, respectively. Lower is better.}\label{tab:e2l}
\end{table}

\paragraph{Comparison with Lagrangian baseline.}

We compare \model{}-DA with GNS~\cite{sanchez2020gns}, a native Lagrangian model trained from
particle trajectories, in Table~\ref{tab:dam2d_gns_tlo}. This comparison
examines whether latent dynamics learned from Eulerian fields, together with
lightweight decoder adaptation, can support competitive particle rollout.
GNS learns particle interactions directly through trajectory supervision,
whereas \model{}-DA keeps its Eulerian-trained encoder and latent processor
fixed and fine-tunes only the decoder using sparse velocity samples at particle
coordinates, without a trajectory loss.

On DAM2D, \model{}-DA achieves more accurate particle rollout with lower
computational overhead than GNS. This advantage comes from the different
representations used for dynamics modeling: GNS explicitly propagates
information among particles through graph message passing, whose cost grows
with particle interactions, while \model{} evolves a compact latent flow
representation and only decodes velocities at queried locations. These results show that an Eulerian latent representation can recover accurate
particle dynamics without the cost of explicit particle interaction modeling.

\begin{table}[t]
\centering
\footnotesize
\setlength{\tabcolsep}{6pt}
\begin{tabular}{lrrrr}
\toprule
Method &
Ref $\downarrow$ &
Path $\downarrow$ &
Mem. $\downarrow$ &
Time $\downarrow$ \\
\midrule
\textbf{TLO-DA}
& \textbf{2.39E-03}
& \textbf{3.72E-03}
& \textbf{35.94}
& \textbf{7.26} \\
GNS
& 4.19E-02
& 2.50E-02
& 47.99
& 8.68 \\
\bottomrule
\end{tabular}
\caption{
Comparison with a native Lagrangian model on DAM2D particle rollout.
TLO and GNS contain 144K and 146K parameters, respectively.
Memory and runtime are reported in MiB and ms/step, respectively.
}\label{tab:dam2d_gns_tlo}
\end{table}

\subsection{Beyond-Grid Lagrangian Rollout}\label{sec:beyond_grid_rollout}

We further evaluate \model{} on Lagrangian rollouts beyond the fixed Eulerian
grid support. Starting from the same latent flow state, we compare two
readout strategies: \model{}-Interp obtains particle velocities by
interpolating the predicted Eulerian field, whereas \model{}-Direct evaluates
the coordinate-conditioned decoder at particle locations.

Within the grid support, the two strategies achieve comparable accuracy. Once particles leave the grid domain, however, standard interpolation is no longer defined without an additional extrapolation rule, while \model{}-Direct can still be evaluated at the off-grid particle coordinates. \model{}-DA, which fine-tunes only the decoder using sparse particle-velocity supervision, further reduces the off-grid rollout error. These results highlight the advantage of decoupling latent flow evolution from coordinate-dependent readout: Lagrangian rollout is not restricted to the spatial support of the Eulerian output grid.

\begin{table}[t]
    \centering

    \label{tab:decoder_adaptation}
    \footnotesize
    \setlength{\tabcolsep}{1pt}
    \renewcommand{\arraystretch}{1.05}
    \begin{tabular}{@{}lccccc@{}}
        \toprule
        & \multicolumn{2}{c}{\shortstack{In-grid\\10 steps}}
        & \multicolumn{2}{c}{\shortstack{Outside\\20 steps}}
        & \shortstack{Full\\20 steps} \\
        \cmidrule(lr){2-3}
        \cmidrule(lr){4-5}
        \cmidrule(l){6-6}
        Method & Ref & Path & Ref & Path & Path \\
        \midrule
        \model{}-Interp
        & \textbf{2.89E-04}
        & \textbf{1.63E-04}
        & -- & -- & -- \\
        \model{}-Direct
        & 3.22E-04
        & 1.89E-04
        & 1.35E-01
        & 6.05E-02
        & 4.48E-03 \\
        \model{}-DA
        & 3.22E-04
        & 1.89E-04
        & \textbf{3.82E-02}
        & \textbf{3.51E-02}
        & \textbf{2.51E-03} \\
        \bottomrule
    \end{tabular}
    \caption{Direct decoding and grid interpolation on DAM2D. A dash indicates that interpolation
    cannot continue after grid exit.}
\end{table}

\begin{figure}[t]
    \centering
    \includegraphics[width=1.05\linewidth]{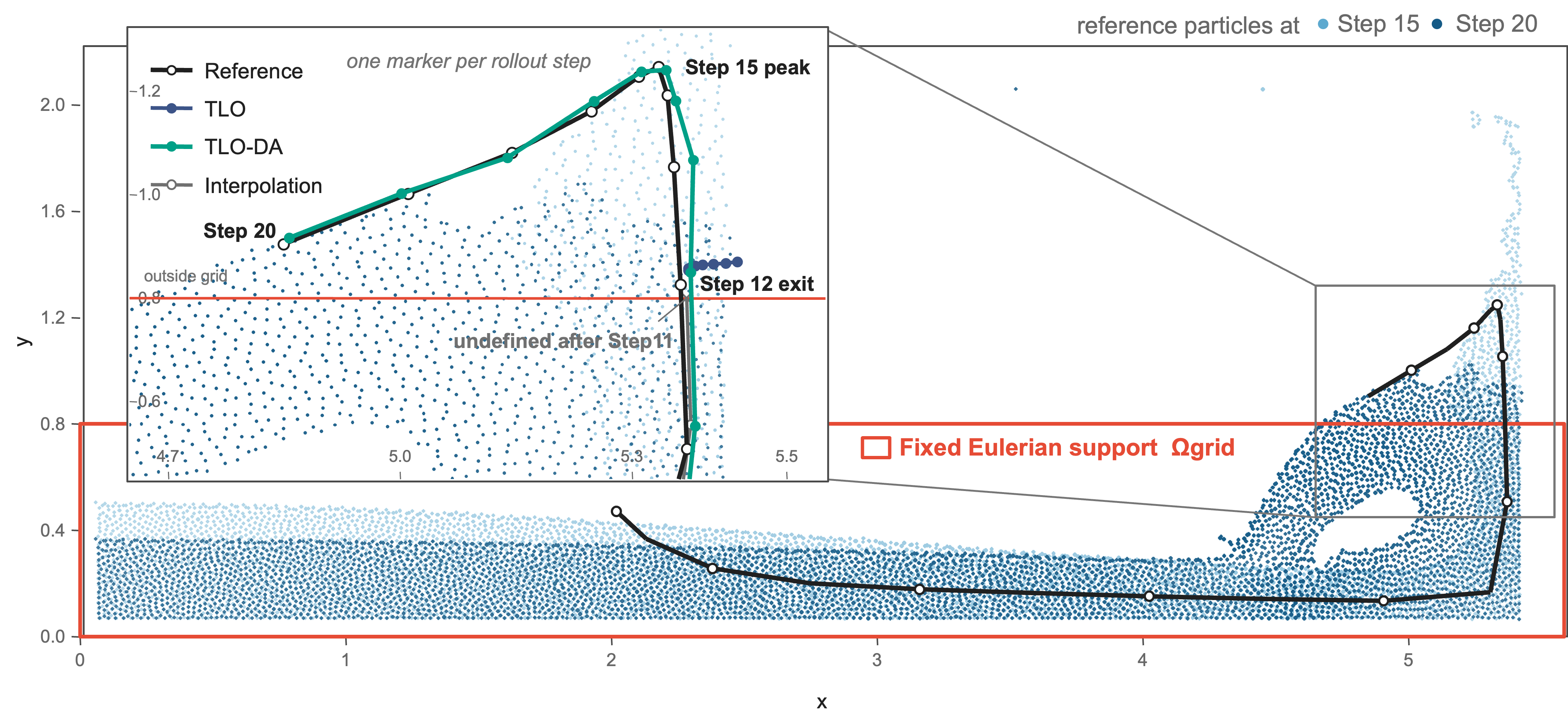}
    \caption{Closed-loop rollout beyond the fixed Eulerian grid support on DAM2D. Light- and
    dark-blue points show reference particles at Steps 15 and 20,
    respectively. The inset follows a particle crossing
    $\Omega_{\mathrm{grid}}$: grid interpolation becomes undefined after the
    exit, whereas direct decoding continues the rollout.}
    \label{fig:out_off_grid}
\end{figure}

\section{Conclusion}

To answer whether flow dynamics learned only from fixed-grid Eulerian data can support closed-loop Lagrangian particle rollout, we formulated zero-shot Eulerian-to-Lagrangian generalization and developed \model{}. By separating query-independent latent dynamics from coordinate-conditioned decoding, \model{} supports both fixed-grid field prediction and rollout at recursively evolving particle locations. Across five field-prediction and three particle-rollout benchmarks, \model{} achieves strong Eulerian accuracy and lower rollout errors than the evaluated baselines; sparse decoder adaptation further improves accuracy and outperforms GNS on DAM2D. The decoder can also be queried after particles leave the fixed-grid domain, whereas interpolation requires an additional extrapolation rule. Overall, these results show that fixed-grid supervision can learn transferable flow dynamics beyond fixed-grid inference, while closed-loop moving queries provide a stringent test of this transferability.

\newpage
\bibliography{arxiv}

\end{document}